\pdfoutput=1

\documentclass[11pt]{article}

\usepackage[preprint]{acl}

\usepackage{times}
\usepackage{latexsym}
\usepackage{comment}
\usepackage{stfloats}
\usepackage[T1]{fontenc}
\usepackage{arydshln}

\renewcommand{\arraystretch}{1.3}

\usepackage{multirow}

\usepackage[utf8]{inputenc}
\usepackage{subfig}
\usepackage{microtype}
\usepackage{float}
\usepackage{inconsolata}

\usepackage{graphicx}

\title{Large Language Model Data Generation for Enhanced Intent Recognition in German Speech}

\author{Theresa Pekarek Rosin \and Burak Can Kaplan \and Stefan Wermter \\
         University of Hamburg - Knowledge Technology\\
        Vogt-Koelln-Strasse 30, 22527 Hamburg - Germany\\
        \url{www.knowledge-technology.info}
         \\
         \small{
    \textbf{Correspondence:} \href{mailto:email@domain}{theresa.pekarek-rosin@uni-hamburg.de}}}

\begin{document}
\maketitle
\begin{abstract}
Intent recognition (IR) for speech commands is essential for artificial intelligence (AI) assistant systems; however, most existing approaches are limited to short commands and are predominantly developed for English.  This paper addresses these limitations by focusing on IR from speech by elderly German speakers. We propose a novel approach that combines an adapted Whisper ASR model, fine-tuned on elderly German speech (SVC-de), with Transformer-based language models trained on synthetic text datasets generated by three well-known large language models (LLMs): LeoLM, Llama3, and ChatGPT. To evaluate the robustness of our approach, we generate synthetic speech with a text-to-speech model and conduct extensive cross-dataset testing. Our results show that synthetic LLM-generated data significantly boosts classification performance and robustness to different speaking styles and unseen vocabulary. Notably, we find that \mbox{LeoLM}, a smaller, domain-specific 13B LLM, surpasses the much larger ChatGPT (175B) in dataset quality for German intent recognition. Our approach demonstrates that generative AI can effectively bridge data gaps in low-resource domains. We provide detailed documentation of our data generation and training process to ensure transparency and reproducibility.

\end{abstract}

\section{Introduction} 
Speech command recognition is essential for natural interaction with artificial agents in everyday life, especially for elderly or handicapped users \citep{Fronemann2021}. Commercial solutions have seen vast improvements and increased usage over the last few years. However, their cloud-based models not only struggle with speech from user groups that would benefit from a voice assistant the most \citep{MoroVelazquez2019,Ngueajio2022}, but they also introduce privacy issues because the processing of speech input usually does not happen locally.

Additionally, traditional speech command recognition is usually restricted to short phrases and words \citep{speechcommandsv2}, which requires the user to adapt to the system and change their way of speaking. That is usually not as intuitive for older people who might not have the same experience with recent technologies as younger users \citep{pekarekrosin2024:framework}. Therefore, we argue that to create a more natural interaction with speech-based assistant systems, we need to move away from one-word command recognition to unconstrained utterance-based intent recognition. 

\begin{figure}[t!]
\centering
  \includegraphics[width=\columnwidth]{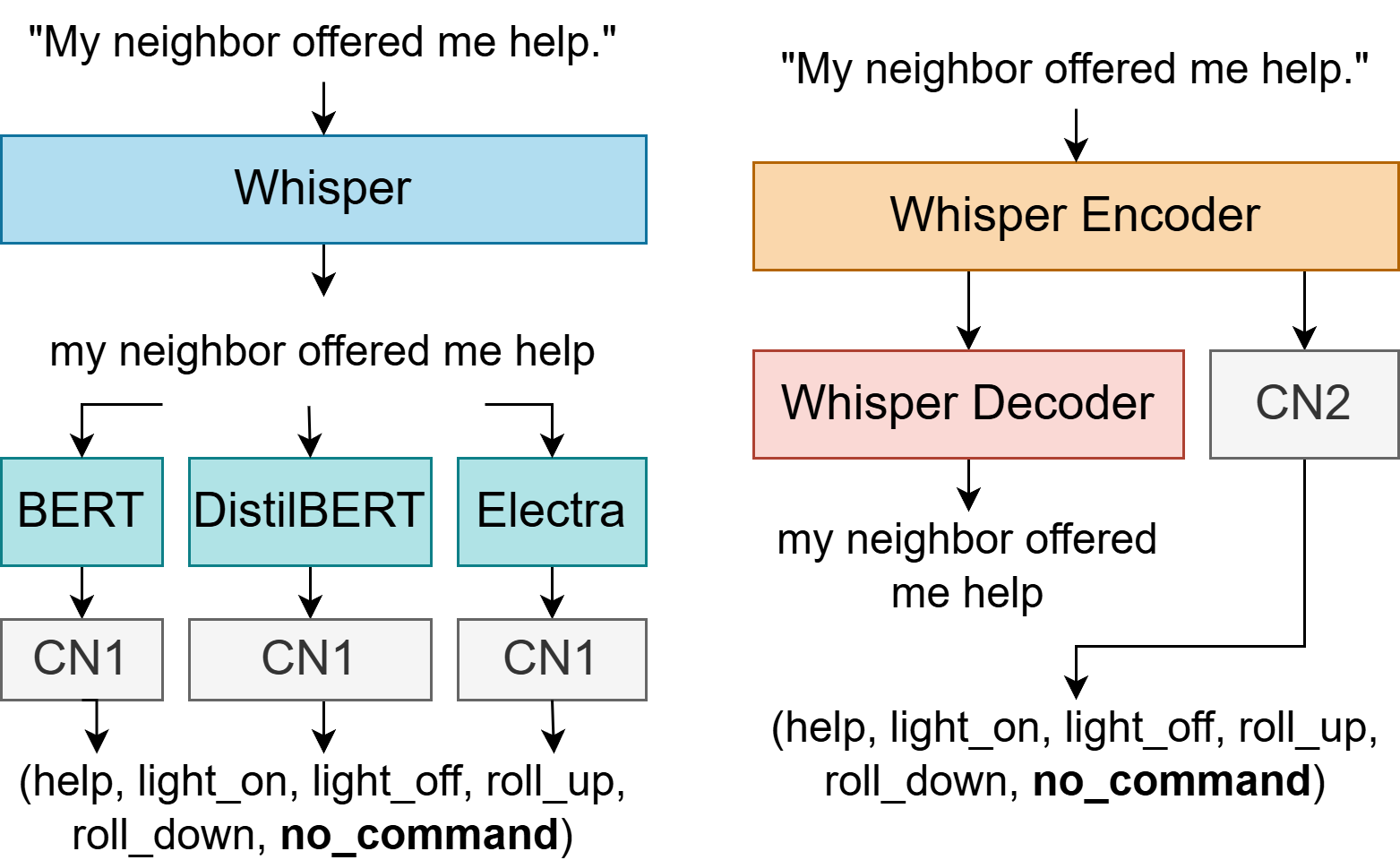}
  \caption{An overview of our model setup (left) and a traditional baseline (right). The speech is transcribed with a Whisper ASR model, the transcript is then classified with a transformer-based LM (BERT, DistilBERT, Electra) trained on a generated text dataset and a one-layer classification network (CN1). For the baseline, we use the output of the Whisper encoder to classify the intent with a two-layer classification network (CN2).}
  \label{fig:approach}
\end{figure}

However, the lack of large-scale datasets for languages besides English makes it difficult to implement such a change by retraining open-source speech models for intent recognition on different domains (e.g., elderly voices).
Collecting new speech data on a larger scale for any speech recognition or classification task is not only an incredible effort for researchers and participants but also ultimately inefficient, since the process would need to be repeated every time new functionalities become available. This issue is amplified by the amount of training data that Transformer-based, state-of-the-art models require to be trained in their entirety.

We suggest two potential solutions to these issues: 1) through the use of layer-specific fine-tuning \citep{Shor2019,PekarekRosin2023} pretrained foundation models can be adapted to other domains with small amounts of speech data while preserving existing knowledge, 2) through the generation of additional domain-focused data with large language models (LLMs), we can increase the generalization abilities of the model to different linguistic patterns.

We combine a state-of-the-art automatic speech recognition (ASR) model, Whisper~\citep{Radford2022}, with three transformer-based pretrained language models, BERT \citep{Devlin2019}, DistilBERT \citep{Sanh2019}, and Electra \citep{Clark2020} for German speech intent recognition. We utilize three LLMs to generate additional training data for these language models, and we synthesize speech samples based on this data with a text-to-speech model to evaluate our approach. 

\section{Related Work} 
\label{sec:relatedwork}

Speech command recognition typically involves accurately identifying words or short phrases as commands linked to functionalities in an underlying system. This classification can be performed on either the audio feature representations or the transcript produced by the ASR model. In contrast, intent recognition requires a larger context, such as complete sentences, to discern user intent. For English-language tasks, the Speech Command Dataset \citep{speechcommandsv2} is a commonly used resource for training and evaluating models in speech command recognition or keyword spotting.

Diverse strategies have been proposed for speech command recognition, with a recent approach by \citet{Sadovsky2023} exploring the use of spiking neural networks for this task, and achieving a maximum accuracy of 72\% on the Speech Command Dataset. Other baselines mentioned in their paper reach 79\% accuracy using Gated Recurrent Units.

\citet{Berg2021} propose the Keyword Transformer model. They train their model on the Speech Command Dataset for approximately 100 episodes and achieve between 97.49\% and 98.56\% accuracy using knowledge distillation with only minimal improvements over the baseline.

In low-resource settings, \citet{KumarNayak20203} explore speech command recognition in the Kui language using a small dataset of 7,090 utterances covering just seven words. Their convolutional neural network (CNN) baseline was the only model to exceed 90\% accuracy after 300 training episodes, highlighting the challenges of limited data. Similarly, \citet{hernandez2024nluforsc} investigate intent recognition in Spanish and Nahuatl by training models on a manually collected dataset of 383 natural language navigation commands. Their results show that transformer-based models outperform traditional baselines, especially for longer and more complex utterances.

Despite these advances, most approaches often rely on extensive training, large-scale datasets, or costly real-world data collection. Notably, the lack of recent work examining this research topic for German speech suggests that the absence of datasets deters research in that direction. 

LLMs have emerged as powerful tools for synthetic dataset generation~\citep{kaplan2025largelanguagemodelsgenerate}, enabling the creation of task-specific training data and offering a solution to data scarcity. Additionally, their capabilities in common-sense reasoning~\cite{li_systematic_2022} and their strong performance in natural language understanding make them well-suited for tasks such as intent recognition. Transformer-based language models have shown value in both data augmentation for intent classification~\citep{kumar-etal-2020-data} and in intent classification itself~\citep{chen2019bert}. 

Additionally, the use of more powerful LLMs for intent recognition has increased significantly \citep{10446132,dzeparoska2024intent,wang2024beyond}, and LLMs have been widely studied for their ability to assist people through multimodal applications with speech~\citep{10446224,padmanabha2024voicepilot}. 
However, the majority of the work has been conducted in English, leaving considerable room for exploration in other languages. While some recent work explores German-language applications of LLMs~\citep{irrgang-etal-2024-features,volk-etal-2024-llm-based}, they have primarily focused on other tasks. The release of German LLMs, such as LeoLM~\footnote{\url{https://laion.ai/blog/leo-lm/}}, has demonstrated the potential of LLMs for the German language and motivates our work, which explores the use of LLM-generated datasets to create robust intent recognition systems for German speakers.

\section{Methodology}
\subsection{Senior Voice Commands Dataset}
The German Senior Voice Commands (SVC-de) is a dataset collected by \citet{PekarekRosin2023} for the development of speech-based interaction with a home assistant system for German senior citizens. The dataset contains recordings from 30 native German speakers (21 female, 9 male) between the ages of 50 and 99, of 52 sentence-based speech commands. Per speaker, approximately 6-7 minutes of audio from two different microphones is available, which leads to a total of 3 hours and 9 minutes of speech data. The recorded sentences can be separated into 6 classes: "help", "light\_on", "light\_off", "roll\_up", "roll\_down", and "no\_command". The class "no\_command" essentially serves to catch common false positives in command classification to decrease wrong triggers, while keeping the interaction natural without the involvement of wake-words or restriction of the user's speech.

\subsection{Intent Recognition Dataset}
\label{subsec:IR}
For our intent recognition task, we generated sentences for the same six classes, with three different large language models (LLMs): LeoLM-13b\footnotemark[1], Llama3-8b\footnote{\url{https://huggingface.co/meta-llama/Llama-3.1-8B}}, and ChatGPT by OpenAI. Llama3 is a well-known recent option for local LLMs, while LeoLM was developed specifically for the German language. 
We included ChatGPT as an assumed upper baseline, since it outperforms most LLMs on a large number of tasks, even though its lack of transparency makes results hard to reproduce. We generated approximately 2500 samples for each LLM.

\begin{figure}[t]
\centering
  \includegraphics[width=\columnwidth]{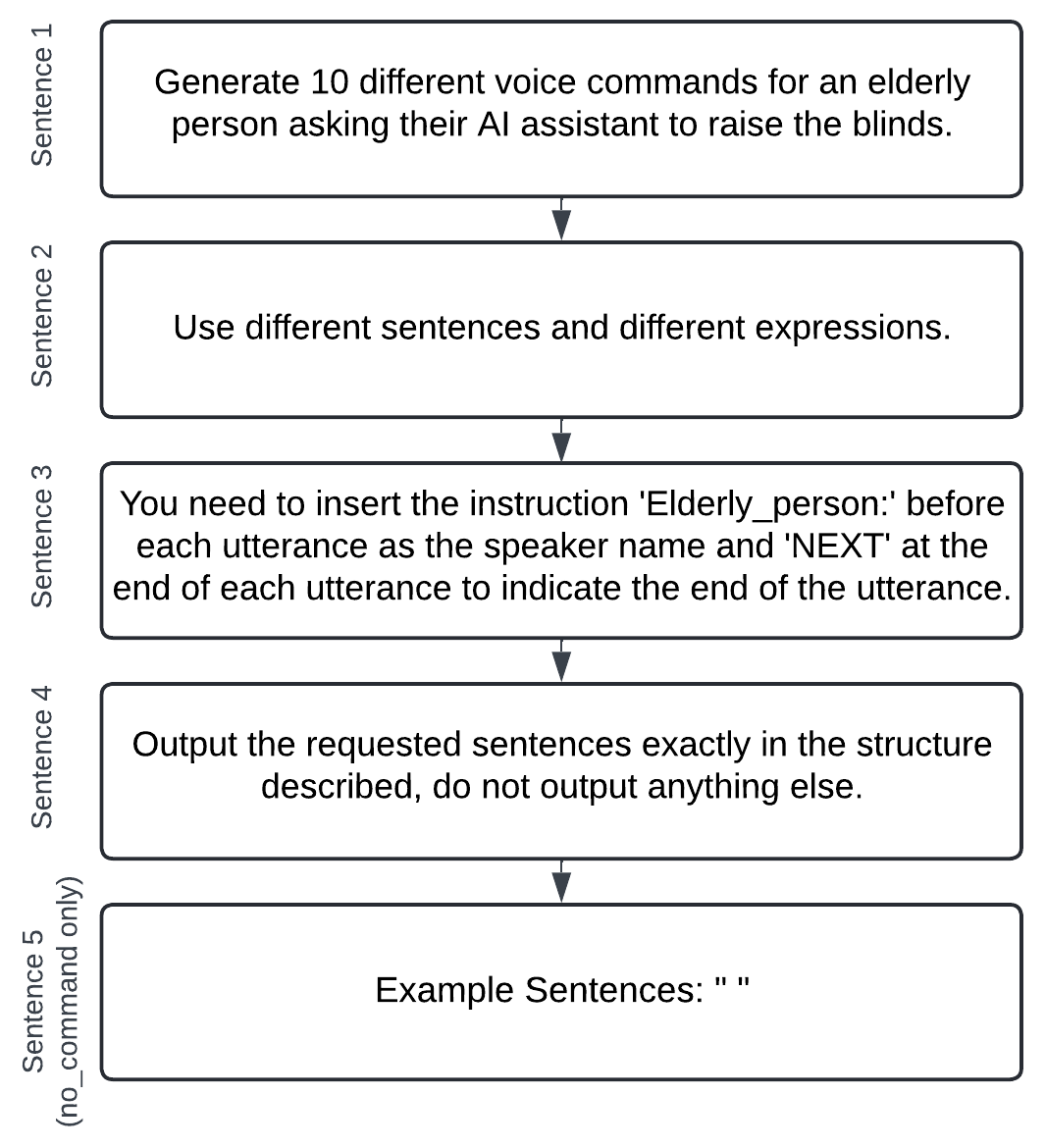}
  \caption{An example for the structure of our prompts. The original prompts are in German and can be found in Appendix~\ref{sec:appendix} in Table~\ref{tab:prompts}.}
  \label{fig:prompt}
\end{figure}

\subsubsection{Prompt Engineering}

We tailored specific prompts for each label to maintain consistency and aimed to keep these prompts as similar as possible. Figure~\ref{fig:prompt} illustrates a general example of the prompt structure. Every prompt, except for the "no\_command" label, is comprised of four sentences, each targeting different sub-tasks. The initial sentence typically clearly outlines the scenario and initiates the task generation. The second sentence aims to enhance diversity by suggesting the LLM should incorporate various situations. The third sentence helps with the structuring of the outputs, making it easier to parse them for data collection. The parsing script is employed post-generation and utilizes the shown keywords to extract the necessary data. Moreover, those keywords provide additional clues and structure to the LLM by clearly indicating the start and end points of the sentences. The final sentence, inspired by \citet{10111523}, is designed specifically to minimize unwanted outputs that usually happen due to the nature of the pre-training of LLMs and their initial underlying prompts. After parsing, each dataset is checked manually to remove sentences with grammatical issues, nonsense content, or ones unrelated to the command they were labeled as.

The "no\_command" label required more prompt engineering than the others, since the concept of false positive sentences does not seem to be easily graspable for LLMs. It often caused the LLM to hallucinate or generate data better suited to other labels. However, we managed to address this issue by employing a few-shot prompting approach with additional example sentences to enrich the existing prompt structure. To further enhance the diversity of the responses, we split the "help" label into two distinct prompts. In addition to the standard prompt, we also requested short calls for help (1-2 words) to account for emergencies, where the user might not be able to speak in full sentences. Additionally, for the "no\_command" label, we tailored three different prompts that correspond to false positive sentences for "help", "light\_on" or "light\_off", and "roll\_up" or "roll\_down", respectively. A few examples of false positive sentences would be: "My assistant turns on the lights as soon as it gets dark.", for control of the lights, "Every morning I open the blinds.", for control of the blinds, and "I should call my doctor later.", for help.

\begin{table}[t]
\centering
\begin{tabular}{llllll}
\hline
\textbf{seed} & \textbf{top\_p} & \textbf{top\_k} & \textbf{RP} & \textbf{typ. p} & \textbf{temp.}\\
\multicolumn{1}{c}{0} & \multicolumn{1}{c}{1} & \multicolumn{1}{c}{10000} & \multicolumn{1}{c}{1}       & \multicolumn{1}{c}{0.995} & \multicolumn{1}{c}{0.7}\\ \hline
\end{tabular}
\caption{The parameters used for the local LLMs. RP: Repetition Penalty}
\label{tab:parametertable}
\end{table}

\begin{table*}[ht]
  \centering
  \begin{tabular}{llrrrr}
    \hline
     & \textbf{\underline{train/test}}      & \underline{ChatGPT} & \underline{LeoLM}  & \underline{Llama3}& \underline{\textbf{Combined}}\\
    &ChatGPT & $95.59\pm0.90$& $79.43\pm1.79$& $71.00\pm1.40$& $81.83\pm1.09$\\
    \textbf{\textit{BERT}} &LeoLM & $82.91\pm2.06$& $94.35\pm0.76$& $76.75\pm1.45$& $\mathbf{84.26\pm0.40}$\\
    &Llama3& $80.00\pm2.68$& $76.31\pm1.00$& $92.01\pm1.23$& $82.41\pm0.44$\\ 
    &\textbf{Combined} & $98.16\pm0.92$& $97.43\pm0.55$& $95.40\pm0.76$& $96.98\pm0.24$\\
    \hline 
     & \textbf{\underline{train/test}}      & \underline{ChatGPT} & \underline{LeoLM}  & \underline{Llama3}& \underline{\textbf{Combined}}\\
    &ChatGPT & $92.99\pm1.08$
& $78.29\pm2.17$& $71.75\pm1.58$& $80.25\pm0.69$\\
    \textbf{\textit{DistilBERT}} &LeoLM & $81.19\pm1.44$
& $93.71\pm0.63$& $70.32\pm1.42$& $\mathbf{82.58\pm0.61}$\\
    &Llama3& $76.76\pm2.76$
& $74.59\pm1.38$& $89.25\pm1.40$& $79.67\pm0.40$\\ 
    &\textbf{Combined} & $97.70\pm0.37$& $96.89\pm0.55$& $94.40\pm0.61$& $96.51\pm0.36$\\
    \hline 
    & \textbf{\underline{train/test}}      & \underline{ChatGPT} & \underline{LeoLM}  & \underline{Llama3}& \underline{\textbf{Combined}}\\
    &ChatGPT & $84.80\pm1.79$& $68.89\pm2.33$& $67.71\pm1.81$& $72.86\pm1.03$\\
    \textbf{\textit{Electra}} &LeoLM & $72.05\pm2.32$& $87.24\pm1.07$& $66.31\pm1.72$& $\mathbf{75.06\pm0.83}$\\
    &Llama3& $66.43\pm1.14$& $71.04\pm1.09$& $81.94\pm2.92$& $72.35\pm1.23$\\ 
    &\textbf{Combined} & $94.88\pm1.01$& $94.35\pm0.81$& $93.85\pm0.67$& $94.33\pm0.36$\\
    \hline
  \end{tabular}
  \caption{\label{tab:tranformer_acc}
    Results for intent recognition from text with all transformer models on all synthetic text datasets. Results are averaged over 5 runs, each model was trained for 5 epochs. We show the mean accuracy (\%) on the test dataset, averaged over 5 checkpoints, and the standard deviation. 
  }
\end{table*}

\subsubsection{LLM Parameters}

We utilized the local LLMs, LeoLM and Llama3, by integrating the TextgenWebUI Chat API~\footnote{\url{https://github.com/oobabooga/text-generation-webui}} and Ollama\footnote{\url{https://ollama.com/}}, which offer similar functionalities as \mbox{OpenAI}'s Chat API, but also allow access to the full range of LLM parameters. 
Table~\ref{tab:parametertable} contains all the parameters we used for the dataset generation for the local LLMs. The seeds were initialized randomly with $0$ and then selected from the range $(0, 2^{35})$. They remained fixed during the generation to ensure the generated data would be diverse and reproducible. In our specific setup, the temperature setting was not as critical due to the stabilized seed; thus, we maintained it at the default value of $0.7$. We selected high values for $top\_ p$, $top\_k$, and $typical\_p$ to enrich the context diversity and expand the array of potential utterances. Additionally, we reduced the repetition penalty ($RP$) to prevent constriction of the LLM with the tokens it previously generated. 

\subsubsection{Speech Synthesis}
Since our model is based on speech input, we utilize XTTS-v2\footnote{\url{https://huggingface.co/coqui/XTTS-v2}}, which is a multilingual text-to-speech model that generates high-quality speech, in its German language configuration.
XTTS-v2 is an extension of the XTTS model by \citet{casanova24_interspeech}, which builds on the Tortoise model \citep{betker2023betterspeechsynthesisscaling} to enable multilingual training, faster inference, and voice cloning.

We randomly select four speakers (two male, two female) between 70 and 80 years of age from the Common Voice DE 10.0 dataset \citep{Ardila2019}, and create short audio files as a reference for the generation of synthetic speech.
We generate audio from these reference files for each LLM-generated text dataset and use the resulting synthetic speech for the complete evaluation of our approach.

\subsection{Architecture}

The task of intent recognition requires reliable speech recognition, not only to enable the use of language models to perform classification on the transcript but also to increase the transparency of the entire model in case of misclassification. Considering these conditions, we use the state-of-the-art Whisper model developed by \citet{Radford2022} in our setup. Whisper follows an encoder-decoder structure, which allows us to neatly utilize the same model for both the baseline and our approach. Specifically, we use a pretrained Whisper-small model that has already been fine-tuned for the German language\footnote{\url{https://huggingface.co/bofenghuang/whisper-small-cv11-german}} and adapt it to the domain of elderly German speech with the SVC-de dataset, following the approach by \citet{PekarekRosin2023}. We continually train only the encoder part of the architecture with Experience Replay \citep{Rolnick2018} on 10\% of the Common Voice DE 10.0 dataset, to avoid overfitting the model on the limited vocabulary of the SVC-de dataset.

In our approach, we combine the domain-adapted Whisper model with a Transformer language model (LM) trained on synthetic text data to perform the task of intent recognition on the transcript provided by the Whisper model (Figure~\ref{fig:approach}). We utilize three pretrained Transformer LMs: BERT \citep{Devlin2019}, DistilBERT \citep{Sanh2019}, and Electra \citep{Clark2020}. BERT was one of the first transformer-based LMs that allowed the adaptation to specific tasks without retraining the entire model, due to the pretrained bi-directional representations in each layer. This property makes BERT and other models like it uniquely suited for low-resource tasks. Additionally, these models have proven reliable on a variety of tasks, and pretrained German versions are readily available in the Hugging Face model repository\footnote{\url{https://huggingface.co/}}.\\ 
DistilBERT reduces BERT's size by 40\% through knowledge distillation, and Electra replaces BERT's masked language modeling task for pre-training with a more sample-efficient one. Since the application context of our approach is real-time interaction with the user, and we want to create a model that could be used locally without access to GPU resources, we utilize these comparatively smaller Transformer models instead of LLMs.

\subsection{Experimental Setup}
In our experimental setup, we take a pretrained BERT (bert-base-german-cased), DistilBERT (distilbert-base-german-cased), and Electra (german-nlp-group/electra-base-german-uncased) model from the Hugging Face model repository\footnotemark[7]. We equip each model with a fully connected output layer for classification and keep the pretrained model frozen. We train the models with a learning rate of 3e-4 and a dropout of $0.1$ over 5 epochs, on the ChatGPT-, LeoLM-, and Llama3-generated text datasets, with a split of 70-20-10 for training, validation, and testing. We selected these hyperparameters empirically, since the models converged after approximately 4 epochs.

Afterward, we perform a cross-evaluation between datasets and also evaluate each model on a combination of all generated datasets, to examine whether there are noticeable differences in quality within the LLM-generated data. This also allows us to assess whether there are LLM-specific patterns in the data and if one of the datasets allows a higher level of generalization to unseen sentences.
 
We then combine the domain-adapted Whisper-small model with the trained LMs and evaluate our architecture on the generated speech datasets. The goal is to compare its performance with the accuracy achieved by the text transformers. We also examine the word error rate (WER) and character error rate (CER) of Whisper on the generated speech datasets and compare them with its performance on SVC-de, to assess the quality of the synthetic speech.

As a baseline, we use the output of the encoder part of the Whisper model to train a two-layer classification network for intent recognition on the SVC-de dataset only. This baseline represents traditional approaches and their limitations in low-resource domains.

\section{Results}

We train and evaluate BERT, DistilBERT, and Electra on each synthetic dataset to examine how well they generalize to different semantic structures and syntax. As seen in Table~\ref{tab:tranformer_acc}, the transformer models trained with data generated by LeoLM show the best generalization abilities across all datasets. While BERT and DistilBERT perform similarly, Electra's accuracy is on average 10\% lower. We also trained the LMs on the SVC-de transcripts and found that the results of the evaluation matched the baseline results in Table~\ref{tab:synthspeech}, with low accuracies (35-40\%) across all synthetic datasets. 

\begin{figure*}[ht]
  \includegraphics[width=1\linewidth]{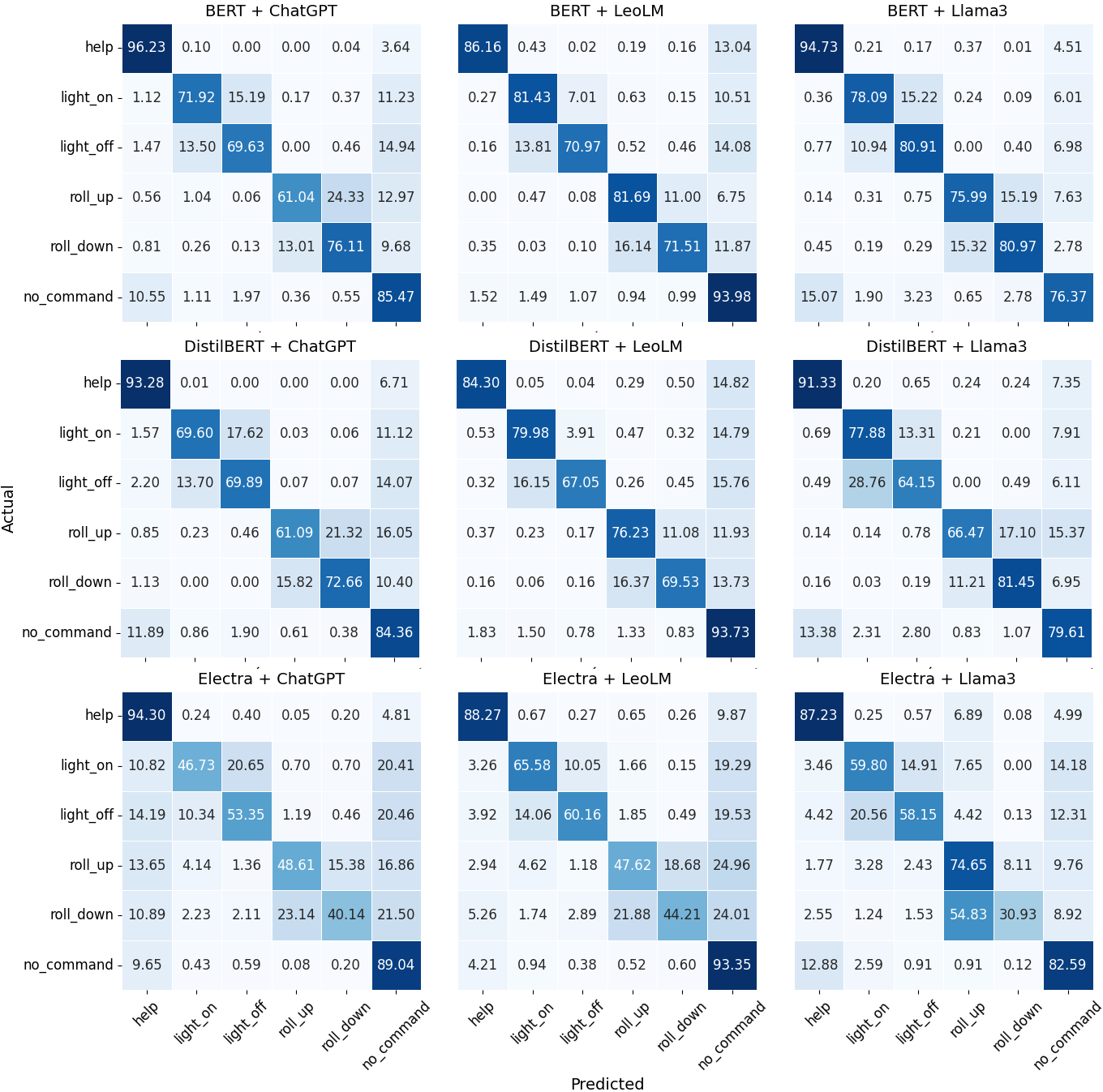}
  \caption{Confusion matrices for all LM+dataset variations, evaluated on the combination of all generated datasets.}
  \label{fig:lm_heatmap}
\end{figure*}

The confusion matrices in Figure~\ref{fig:lm_heatmap} show that models trained on the LeoLM data generally show a more stable performance, even for unseen samples generated by other models. All models exhibit some level of confusion with the differentiation of "light\_on" and "roll\_up" from their counterparts and "no\_command". However, BERT outperforms both Electra and DistilBERT. Electra seems to have more issues overall with the separation of different classes, which is also reflected in the previously discussed lower accuracies. A notable, positive result is the fact that all models across all datasets seem to be able to distinguish real calls for help reliably from the false positives introduced in the data for "no\_command".

We continue the evaluation of our approach and the baseline with synthetic speech, as discussed in Section~\ref{subsec:IR}. As can be seen in Table~\ref{tab:wer}, the quality of the generated speech is within an acceptable range, considering the domain-adapted ASR model was not trained on the synthetic speech. We observe word error rates (WERs) of 12.14\% for ChatGPT, 15.01\% for LeoLM, and 9.56\% for Llama3. Meanwhile, the real speech from the SVC-de dataset achieves a WER as low as 5.65\%, which is in line with the results by \citet{PekarekRosin2023}.

\begin{table}
  \centering
  \begin{tabular}{lcc}
    \hline
    \textbf{Dataset} & \textbf{WER (\%) $\downarrow$} & \textbf{CER (\%) $\downarrow$}\\
    \hline
    \textbf{ChatGPT} & 12.14 & 6.16\\
    \textbf{LeoLM} & 15.01 & 8.41\\
    \textbf{Llama3} & 9.56 & 4.94\\ \hline
    \textbf{SVC-de} & 5.65  & 5.42\\ \hline
  \end{tabular}
  \caption{The word error rate (WER) and the character error rate (CER) of the domain-adapted Whisper-small on the real-world SVC-de dataset and the synthetic speech datasets.}
  \label{tab:wer}
\end{table}

The results in Table~\ref{tab:synthspeech} show that pre-training the LMs on the data generated by the LLMs increases the overall models' robustness against unseen sentences. This is supported by the significantly lower performance of the baseline on the synthetic data, with around a 50\% difference in accuracy. The LeoLM data proves to be especially useful for pre-training, with Whisper+DistilBERT(LeoLM) achieving the highest accuracy with 83.01\% on SVC-de compared to the baseline result of 95.05\%. It even outperforms the models trained on a combination of the synthetic text data. The same patterns that can be observed in the evaluation of the text data (Table~\ref{tab:tranformer_acc}) are present here as well, with a performance drop for Electra-based models. Overall, we observe that adding a language model trained on a supplementary synthetic text dataset vastly improves generalization to new data, compared to the baseline results.

\begin{table*}
  \centering
  \begin{tabular}{llrrrr}
    \hline
         & \underline{\textbf{train/test}} & \underline{ChatGPT-s}& \underline{LeoLM-s}& \underline{Llama3-s}& \underline{\textbf{SVC-de}}\\
 
    &ChatGPT
    & $91.49\pm0.95$&  $77.51\pm1.44$& $71.35\pm1.54$& $65.91\pm3.86$\\
    \textbf{\textit{Whisper+BERT}}& LeoLM
    & $78.28\pm0.46$& $89.95\pm0.43$& $72.88\pm1.06$& $\mathbf{73.51\pm2.65}$\\
    & Llama3& $76.04\pm1.11$& $71.99\pm0.88$& $91.09\pm0.70$& $63.44\pm6.63$\\ 
    & \textbf{Combined} & $93.63\pm0.44$& $92.32\pm0.26$& $94.23\pm0.37$& $74.37\pm2.87$\\ \hline
    & ChatGPT
    & $87.50\pm1.60$& $73.93\pm0.83$& $68.11\pm2.23$& $75.58\pm3.80$\\
    \multirow{4}{*}{\raisebox{6.5ex}{\textbf{\textit{Whisper+DistilBERT}}}} & LeoLM
    & $74.29\pm1.37$& $86.69\pm1.42$& $67.43\pm1.80$& $\mathbf{83.01\pm1.54}$\\
    & Llama3& $75.04\pm1.06$& $71.93\pm0.83$& $87.04\pm1.07$& $71.47\pm1.89$\\ 
    & \textbf{Combined} & $94.35\pm0.11$& $92.13\pm0.21$& $92.12\pm0.42$& $72.28\pm1.35$\\ \hline
    & ChatGPT 
    & $74.16\pm1.34$& $65.50\pm0.42$& $64.48\pm0.84$& $61.67\pm0.96$\\
    \multirow{4}{*}{\raisebox{6.5ex}{\textbf{\textit{Whisper+Electra}}}} & LeoLM
    & $65.94\pm0.63$& $80.97\pm0.36$& $63.35\pm0.74$& $\mathbf{71.42\pm2.27}$\\
    & Llama3& $67.16\pm1.01$& $69.05\pm1.64$& $81.76\pm1.73$& $51.52\pm4.53$\\
    & \textbf{Combined} & $87.32\pm0.65$& $86.48\pm0.67$& $90.10\pm0.32$& $68.61\pm1.09$\\ 
    \hline 
    \textbf{\textit{Whisper-Encoder+CN}} & SVC-de& $34.72\pm0.61$& $38.40\pm0.65$& $36.36\pm0.75$& $95.05\pm0.39$\\
    \hline
  \end{tabular}
  \caption{\label{tab:synthspeech}
    Results of the evaluation on the synthetic speech datasets (-s), and on the SVC-de dataset of all combinations of Whisper+LM models, trained on the text datasets generated by ChatGPT, LeoLM, and Llama3. Whisper-Encoder+CN is the baseline trained only on the real speech dataset, SVC-de. We show the mean accuracy (\%), averaged over 5 checkpoints, and the standard deviation. 
  }
\end{table*}

\section{Discussion}
\label{sec:discussion}

In our approach, we equip a Whisper ASR model with a Transformer LM for intent recognition, to allow LLM-generated text data to supplement the small-scale speech dataset available for the task (SVC-de). 
Our results show that training the LM on the synthetic text data increases the overall model's ability to generalize across various semantic structures and syntax, and previously unseen sentences. 

LeoLM, a local LLM specifically created for the German language, outperforms both ChatGPT and Llama3 in terms of generalization to unseen data (Table~\ref{tab:tranformer_acc}) and demonstrates a more stable performance overall (Figure~\ref{fig:lm_heatmap}). This shows the potential of language-specific fine-tuning for local LLMs, which offer more transparency of their parameters and therefore higher control of the output and reproducibility. Notably, models trained only on the Llama3 data do not trail behind the ones trained on LeoLM or ChatGPT, even though Llama3 is a smaller model. For the BERT models in particular, performance differences are minimal, and BERT(Llama3) even outperforms BERT(ChatGPT) on the dataset combination. 

The confusion matrices in Figure~\ref{fig:lm_heatmap} show that the distinction between "help" and the false positives contained in "no\_command" seems to be straightforward for all models, which is great for real-life applications, since calls for help should ideally not be misclassified at all. All models seem to struggle to varying degrees with differentiating between "light\_on" or "light\_off" and "roll\_up" or "roll\_down", which is to be expected due to the high similarity of the generated sentences. The Electra models seem to have more issues with this differentiation overall, especially the model trained on the ChatGPT data. In a real home assistant system, this would ideally be alleviated by introducing context information to help reduce uncertainty, e.g., by checking the state of the lights or blinds.

The word error rates (WERs) of the synthetic speech datasets (Table~\ref{tab:wer}) are comparable to the ones measured by \citet{PekarekRosin2023} for Whisper-small-de on the Common Voice DE test split (11.2\%). This indicates that the quality of the synthetic speech approximates real speech. 
As can be seen in Table~\ref{tab:synthspeech}, the evaluation results on the synthetic speech datasets and SVC-de show that BERT and DistilBERT can keep the classification performance high for unseen speech data. The performance for the speech-based evaluation drops slightly across all models compared to the text-only evaluations, which is expected since ASR models usually create noisy transcriptions (Table~\ref{tab:wer}), and some information might be lost. 
We also observe that supplementing with data from a single LLM (LeoLM) seems to be more beneficial than combining the data from all LLMs for training.

The low performance of the baseline Whisper-Encoder+CN on the synthetic datasets highlights how fine-tuning on small-scale speech data can significantly affect pretrained model generalizability and emphasizes the advantage of our approach. Since the WER of the synthetic speech is comparable to real speech, the limited vocabulary of the real dataset likely explains the baseline's low accuracy. 
Additionally, while SVC-de was used for domain adaptation of the Whisper encoder, none of the Whisper+LM variations were explicitly trained on it for classification. Still, all models achieved above-average accuracy on SVC-de, with Whisper+DistilBERT(LeoLM) outperforming all other model variations and approaching the baseline trained on SVC-de (Whisper-Encoder+CN). 

Finally, all our models require only 5 epochs to be trained sufficiently, which can be done in a fraction of the time needed for other approaches (Section~\ref{sec:relatedwork}), since each training run on the generated text data takes only 1-2 minutes.

\section{Conclusion}

In this paper, we present a novel approach for intent recognition (IR) in the domain of elderly German speakers using a home assistant system. We leverage a pretrained Whisper model and adapt it to the domain through layer-specific fine-tuning and continual learning on the SVC-de dataset. To address the limitations of the dataset, we generate supplementary classification datasets with three large language models (LLMs): LeoLM, Llama3, and ChatGPT. These datasets are then used to train transformer-based language models (LMs) for IR.

Our results show that a pretrained ASR model combined with an LM trained on synthetic text data displays increased robustness to diverse linguistic patterns and unseen vocabulary. Evaluating our models on high-quality synthetic speech shows that they outperform the baseline trained only on real-world data (SVC-de), indicating improved robustness to different speakers. This adaptability is critical for reducing user strain in real-world applications. We find that training Transformer LMs on synthetic text data is more efficient than continued ASR fine-tuning in terms of resources and generalizability of the model. Additionally, our models are reliably able to differentiate calls for help from false alarms, which is essential for a home-assistant system for elderly speakers.

This work is an exploration of LLMs and generative AI for the generation of German speech and language processing datasets. We show that (1) a domain-specific, smaller LLM like LeoLM (13B) can surpass larger models like ChatGPT (175B) in dataset quality, while offering transparency and reproducibility; (2) supplementing an ASR model with a language model trained on synthetic text data can enhance model performance and robustness; and (3) our method offers a fast and efficient way to adapt existing speech systems for IR tasks. 
As such, our approach offers a practical and scalable solution for deploying reliable home-assistant systems in real-world speaker environments.

\section*{Ethical Considerations}
No experiments with human participants or additional recordings were conducted during this research. We chose to use local LLMs as well as ChatGPT for the dataset generation to explore their performance, since local LLMs allow for greater control of their parameters. For the sake of transparency, we have shared every parameter (Table~\ref{tab:parametertable}) and prompt (Appendix~\ref{sec:appendix}) we used. The generated data can be fully reproduced using the same LLMs, prompts, and parameters, provided that the same seed is used.

\section*{Limitations}

In this research, we have encountered a couple of limitations associated with the use of smaller LLMs. It was observed that the LLM may truncate its outputs. To address this issue, we implemented a check during the parsing phase to verify the completeness of each utterance in terms of starting and ending keywords. If incomplete, the parser excludes that utterance, ensuring that the parsed data does not contain missing values. We also observed instances where recurrent utterances appear in more than one output. This issue is resolved by eliminating duplicate utterances during parsing and selecting the more diverse ones during manual filtering.
For the ASR model, we chose one model to simplify the experimental setup, but ideally, we would have examined the performance of different versions of Whisper as well. The approach we follow for domain adaptation does an extensive evaluation with the same dataset, so we did not repeat a similar evaluation.

\section*{Acknowledgments}
The authors gratefully acknowledge funding from Horizon Europe under the MSCA grant agreement No 101072488 (TRAIL), the German BMWK (SIDIMO), and the Study Abroad Graduate Scholarship by the Ministry of National Education of Türkiye.

\bibliography{anthology,custom}

\appendix
\section{Prompts}
\label{sec:appendix}
\begin{table*}[t]
\renewcommand{\arraystretch}{0.95} 
\footnotesize
\centering
\begin{tabular}{c|p{13.9cm}}

\textbf{Label} & \parbox{\linewidth}{\centering \textbf{Prompt}} \\[0.2ex]

\hline\\[-1ex]
\multirow{2}{*}{\raisebox{-15ex}{\centering\rotatebox{90}{help}}} & Generieren Sie 10 verschiedene Sprachbefehle, mit denen ein älterer Mensch seinen KI-Assistenten in gefährlichen Situationen um Hilfe bitten kann, ohne jedes Mal explizit um Hilfe zu bitten. Verwenden Sie verschiedene Sätze und unterschiedliche Gesundheitssituationen. Sie müssen vor jeder Äußerung die Anweisung 'Ältere\_Person:' als Sprechername einfügen und 'NÄCHSTES' am Ende jeder Äußerung, um das Ende der Äußerung zu kennzeichnen. Geben Sie die angeforderten Sätze genau in der beschriebenen Struktur aus, geben Sie nichts anderes aus. \\[0.7ex]
\cdashline{2-2}[0.5pt/2pt]\\[-1.1ex]
& Generieren Sie 10 verschiedene sehr kurze Sprachbefehle, mit denen eine ältere Person ihren KI-Assistenten in gefährlichen Situationen um Hilfe bitten kann, ohne jedes Mal explizit um Hilfe zu bitten. Verwenden Sie verschiedene, kurze Sätze, die aus ein bis zwei Wörtern bestehen und verschiedene Gesundheitssituationen beschreiben. Fügen Sie vor jeder Äußerung die Anweisung 'Ältere\_Person:' als Sprechername ein und am Ende jeder Äußerung 'NÄCHSTES', um das Ende der Äußerung zu markieren. Geben Sie die geforderten Sätze genau in der beschriebenen Struktur aus, geben Sie nichts anderes aus. \\[0.7ex]
\hline\\[-1.2ex]

\multirow{2}{*}{\raisebox{-12ex}{\centering\rotatebox{90}{roll}}} & Generieren Sie 10 verschiedene Sprachbefehle für eine ältere Person, die ihren KI-Assistenten bittet, die Rollläden hochzufahren. Verwenden Sie verschiedene Sätze und unterschiedliche Ausdrücke. Sie müssen vor jeder Äußerung die Anweisung ‘Ältere\_Person:‘ als Sprechername einfügen und ‘NÄCHSTES‘ am Ende jeder Äußerung, um das Ende der Äußerung zu kennzeichnen. Geben Sie die angeforderten Sätze genau in der beschriebenen Struktur aus, geben Sie nichts anderes aus. \\[0.7ex]
\cdashline{2-2}[0.5pt/2pt]\\[-1.1ex]
& Generieren Sie 10 verschiedene Sprachbefehle für eine ältere Person, die ihren KI-Assistenten bittet, die Rollläden herunterzufahren. Verwenden Sie verschiedene Sätze und unterschiedliche Ausdrücke. Sie müssen vor jeder Äußerung die Anweisung 'Ältere\_Person:' als Sprechername einfügen und 'NÄCHSTES' am Ende jeder Äußerung, um das Ende der Äußerung zu kennzeichnen. Geben Sie die angeforderten Sätze genau in der beschriebenen Struktur aus, geben Sie nichts anderes aus. \\[0.7ex]
\hline\\[-1.2ex]

\multirow{2}{*}{\raisebox{-13ex}{\centering\rotatebox{90}{lights}}} & Generieren Sie 10 verschiedene Sprachbefehle für eine ältere Person, die ihren KI-Assistenten bittet, das Licht einzuschalten. Verwenden Sie verschiedene Sätze und unterschiedliche Ausdrücke. Sie müssen vor jeder Äußerung die Anweisung 'Ältere\_Person:' als Sprechername einfügen und 'NÄCHSTES' am Ende jeder Äußerung, um das Ende der Äußerung zu kennzeichnen. Geben Sie die angeforderten Sätze genau in der beschriebenen Struktur aus, geben Sie nichts anderes aus. \\[0.7ex]
\cdashline{2-2}[0.5pt/2pt]\\[-1.1ex]
& Generieren Sie 10 verschiedene Sprachbefehle für eine ältere Person, die ihren KI-Assistenten bittet, das Licht auszuschalten. Verwenden Sie verschiedene Sätze und unterschiedliche Ausdrücke. Sie müssen vor jeder Äußerung die Anweisung 'Ältere\_Person:' als Sprechername einfügen und 'NÄCHSTES' am Ende jeder Äußerung, um das Ende der Äußerung zu kennzeichnen. Geben Sie die angeforderten Sätze genau in der beschriebenen Struktur aus, geben Sie nichts anderes aus. \\[0.7ex]
\hline\\[-1.2ex]

\multirow{3}{*}{\raisebox{-40ex}{\centering\rotatebox{90}{no\_command}}} & Generieren Sie 10 Sätze von einer älteren Person, die von einer Spracherkennung fälschlicherweise als 'Bitte um Hilfe' klassifiziert werden können, aber in Wirklichkeit als 'kein Befehl' für einen KI-Assistenten verwendet werden. Der Assistent benutzt dafür eine Keyword Detection. Verwenden Sie verschiedene Sätze und verschiedene Ausdrücke. Sie müssen vor jeder Äußerung als Sprechernamen die Anweisung 'Ältere\_Person:' und am Ende jeder Äußerung 'NÄCHSTES' einfügen, um das Ende der Äußerung anzuzeigen. Ein paar Beispiele: Ältere\_Person: Kannst du mir bitte helfen, mein Handy zu finden? NÄCHSTES Ältere\_Person: Mein Sohn hat mir gestern mit dem Garten geholfen. NÄCHSTES Ältere\_Person: Manchmal muss ich um Hilfe bitten. NÄCHSTES Ältere\_Person: Diese neuen Geräte sind ohne Hilfe gar nicht zu bedienen. NÄCHSTES Ältere\_Person: Früher konnte ich alles alleine, ohne um Hilfe zu bitten. NÄCHSTES \\[0.7ex]
\cdashline{2-2}[0.5pt/2pt]\\[-1.1ex]
& Generieren Sie 10 Sätze von einer älteren Person, die von einer Spracherkennung fälschlicherweise als 'Rollläden hoch- oder runterfahren' klassifiziert werden können, aber in Wirklichkeit als 'kein Befehl' für einen KI-Assistenten verwendet werden. Der Assistent benutzt dafür eine Keyword Detection. Verwenden Sie verschiedene Sätze und verschiedene Ausdrücke. Sie müssen vor jeder Äußerung als Sprechernamen die Anweisung 'Ältere\_Person:' und am Ende jeder Äußerung 'NÄCHSTES' einfügen, um das Ende der Äußerung anzuzeigen. Ein paar Beispiele: Ältere\_Person: Mein Assistent fährt die Rollläden jeden Abend pünktlich um 18:00 herunter. NÄCHSTES Ältere\_Person: Im Sommer habe ich die Jalousien gerne den ganzen Tag unten. NÄCHSTES Ältere\_Person: Es ist sehr praktisch, dass mein Sprachassistent die Rollläden steuern kann. NÄCHSTES Ältere\_Person: Meine Rollläden sind beim letzten Sturm kaputtgegangen. NÄCHSTES Ältere\_Person: Sobald meine Jalousien oben sind, kann ich meinen Tag beginnen. NÄCHSTES \\[0.7ex]
\cdashline{2-2}[0.5pt/2pt]\\[-1.1ex]
& Generieren Sie 10 Sätze von einer älteren Person, die von einer Spracherkennung fälschlicherweise als 'Licht ein- oder ausschalten' klassifiziert werden können, aber in Wirklichkeit als 'kein Befehl' für einen KI-Assistenten verwendet werden. Der Assistent benutzt dafür eine Keyword Detection. Verwenden Sie verschiedene Sätze und verschiedene Ausdrücke. Sie müssen vor jeder Äußerung als Sprechernamen die Anweisung 'Ältere\_Person:' und am Ende jeder Äußerung 'NÄCHSTES' einfügen, um das Ende der Äußerung anzuzeigen. Ein paar Beispiele: Ältere\_Person: Mein Assistent schaltet mir jeden morgen die Lichter an. NÄCHSTES Ältere\_Person: Gestern hatten wir schon sehr früh kein Licht mehr im Raum. NÄCHSTES Ältere\_Person: Die Tatsache, dass mein Sprachassistent das Licht an- und ausschalten kann ist sehr praktisch. NÄCHSTES Ältere\_Person: Da ist mir ein Licht aufgegangen. NÄCHSTES Ältere\_Person: Manchmal ist es hier ziemlich dunkel ohne Licht. NÄCHSTES \\[0.7ex]
\hline

\end{tabular}
\caption{All the prompts used for the different classes in the datasets.}
\label{tab:prompts}
\end{table*}

\end{document}